\documentclass[twocolumn,letterpaper]{IEEEAerospaceCLS}  

\usepackage{graphicx,import}  
\usepackage{layouts}

\usepackage{booktabs}
\usepackage{adjustbox}
\usepackage{lscape}
\usepackage{array}
\usepackage{amsfonts}
\usepackage{amsmath} 
\usepackage[acronym]{glossaries}
\usepackage{siunitx}
\usepackage[inline]{enumitem}
\usepackage[textsize=scriptsize]{todonotes} 

\usepackage{tabularx}
\usepackage{hyperref}
\hypersetup{colorlinks=true}
\usepackage[capitalise]{cleveref}
\usepackage[justification=justified]{caption}

\newcolumntype{R}[2]{%
    >{\adjustbox{angle=#1,lap=\width-(#2)}\bgroup}%
    l%
    <{\egroup}%
}
\newcommand*\rot{\multicolumn{1}{R{90}{1em}}}
\newcommand*\rott{\multicolumn{1}{R{45}{1em}}}

\newcommand{\ignore}[1]{}  
\newcommand\thefontsize{The current font size is: \f@size pt}

\newcommand{\ie}{i.\,e.,\ }
\newcommand{\etal}{\textit{et al.}}

\makeglossaries

\begin{document}
\title{How Important are Data Augmentations to Close the Domain Gap for Object Detection in Orbit?}

\author{%
Maximilian Ulmer$^{1,2}$\\ 
\and 
Leonard Klüpfel$^{1}$\\
\and 
Maximilian Durner$^{1,3}$\\
\and 
Rudolph Triebel$^{1,2}$\\
\\
\thanks{$^{1}$ Institute of Robotics and Mechatronics, German Aerospace Center (DLR), 82234 Wessling, Germany {\tt\small \texttt{<first>.<second>@dlr.de}}}%
\thanks{$^{2}$Department of Informatics, Karlsruhe Institute of Technology, 76131 Karlsruhe, Germany}
\thanks{$^{3}$Department of Computer Science, Technical University of Munich (TUM), 85748 Garching, Germany}
}

\maketitle

\thispagestyle{plain}
\pagestyle{plain}

\maketitle

\thispagestyle{plain}
\pagestyle{plain}

\newacronym{AP}{AP}{Average Precision}
\newacronym{BCE}{BCE}{Binary Cross Entropy}
\newacronym{CE}{CE}{Cross Entropy}
\newacronym{CNN}{CNN}{convolutional neural network}

\newacronym{DL}{DL}{Deep Learning}
\newacronym{EMA}{EMA}{Exponential Moving Average}
\newacronym{ESA}{ESA}{European Space Agency}

\newacronym{fANOVA}{fANOVA}{functional analysis of variance}
\newacronym{GMM}{GMM}{Gaussian Mixture Model}
\newacronym{GPU}{GPU}{graphics processing unit}
\newacronym{GPUs}{GPUs}{graphics processing units}
\newacronym{GAN}{GAN}{generative adverserial network}

\newacronym{HIL}{HIL}{hardware-in-the-loop}
\newacronym{IoU}{IoU}{Intersection over Union}
\newacronym{mAP}{mAP}{Mean Average Precision}
\newacronym{PnP}{PnP}{Perspective-n-Point}
\newacronym{RoI}{RoI}{Region of Interest}
\newacronym{RANSAC}{RANSAC}{Random Sample Consensus}
\newacronym{SPEC}{SPEC}{Satellite Pose Estimation Challenge}
\newacronym{TPE}{TPE}{Tree-structured Parzen Estimator}
\newacronym{VOC}{VOC}{PASCAL Visual Object Classes}
\newacronym{yolo}{YOLO}{You Only Look Once}

\begin{abstract}
We investigate the efficacy of data augmentations to close the domain gap in spaceborne computer vision, crucial for autonomous operations like on-orbit servicing. 
As the use of computer vision in space increases, challenges such as hostile illumination and low signal-to-noise ratios significantly hinder performance. 
While learning-based algorithms show promising results, their adoption is limited by the need for extensive annotated training data and the domain gap that arises from differences between synthesized and real-world imagery. 
This study explores domain generalization in terms of data augmentations -- classical color and geometric transformations, corruptions, and noise -- to enhance model performance across the domain gap. 
To this end, we conduct an large scale experiment using a hyperparameter optimization pipeline that samples hundreds of different configurations and searches for the best set to bridge the domain gap.
As a reference task, we use 2D object detection and evaluate on the SPEED+ dataset that contains real hardware-in-the-loop satellite images in its test set.
Moreover, we evaluate four popular object detectors, including Mask R-CNN, Faster R-CNN, YOLO-v7, and the open set detector GroundingDINO, and highlight their trade-offs between performance, inference speed, and training time.
Our results underscore the vital role of data augmentations in bridging the domain gap, improving model performance, robustness, and reliability for critical space applications. 
As a result, we propose two novel data augmentations specifically developed to emulate the visual effects observed in orbital imagery. 
We conclude by recommending the most effective augmentations for advancing computer vision in challenging orbital environments. Code for training detectors and hyperparameter search will be made publicly available.
\end{abstract}


\tableofcontents

\section{Introduction}
 \begin{figure}
\includegraphics[width=0.48\textwidth]{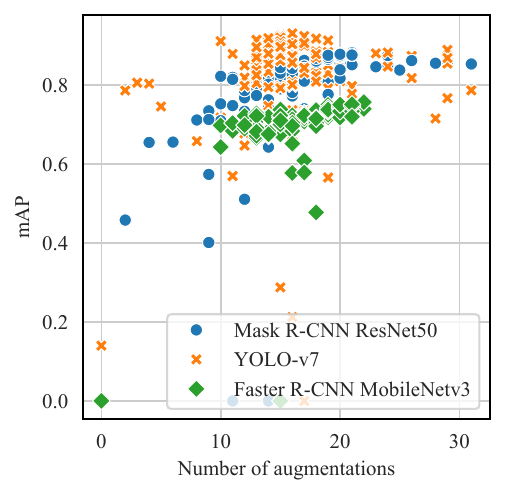}
\caption{Impact of data augmentations on satellite object detection to bridge the domain gap. 
We observe that detectors predict more precise bounding box on real-world images when trained with more augmentation techniques applied during the training on synthetic images.}
\label{fig:n_augs_vs_perf}
\vspace{-0.3em}
\end{figure}
\begin{figure*}[ht!]
    \centering
    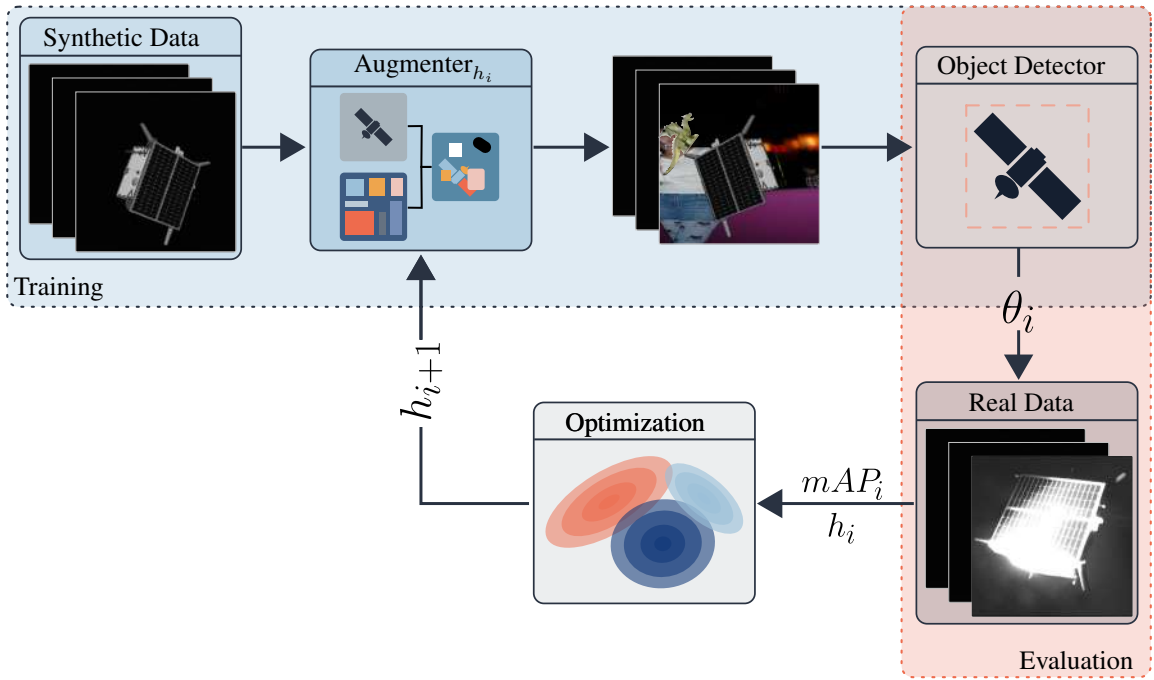
    \caption{
    Overview on our hyperparameter optimization strategy for finding the set of data augmentations yielding the best precision when trained on synthetic and tested on real-world orbital satellite images.
    To this end, we train a detection model with varying augmentations techniques, report an evaluation score that informs the next set of augmentations to experiment with in the consecutive trial.}
    \label{fig:overview}
\end{figure*}

The application of autonomous systems in space is increasing as humans’ presence in orbit expands and the sustainability of the orbital environment as well as its infrastructure is becoming increasingly important. 
Especially for highly complex, autonomous operations, such as on-orbit servicing, accurate and high-performance computer vision is a necessity. 
However, the orbit is an extremely challenging domain for camera-based algorithms due to hostile illumination, high contrast, and low signal-to-noise ratio. 
Such conditions can drastically deteriorate the performance of classical computer vision algorithms due to obscured color and geometric features. 
Learning-based algorithms have recently shown convincing capabilities to deal with such conditions, yet, the space-domain remains hesitant to widely adopt these methods. 
Besides the current lack of relevant hardware, one reason is the vast amounts of annotated training data that learning-based methods require to achieve good performance. 
To this end, recent work focuses on using rendered data for training and transferring the learned model to real data during inference~\cite{Park2021-jr}. 
This divergence between the training and inference data distributions is called the domain gap. 
The effects of the gap are especially drastic in the orbital domain, likely due to the difficulty of rendering tools to achieve a high degree of visual realism.
Spacecrafts are often constructed from materials causing specular reflections that result in visual artifacts, such as a blooming and overexposure. 
Most objects in space are partially covered in a reflective insulation layer that can only be approximated in the 3D model and is challenging to render.  
Similarly, it is very expensive to synthesize the low signal-to-noise ratio and lack of atmospheric diffusion.
These effects result in an especially wide domain gap for orbital imagery, making transferring a model more difficult~\cite{Park2023-qf}. 

Domain generalization is one of the primary approaches to deal with such a domain gap if the target domain is unknown or annotated target data is too costly~\cite{Wang2021-ou}. 
It aims to augment the training procedure to learn domain invariant features or change the training data statistics to reduce the divergence between training and testing distributions.
To achieve the latter, one of the most effective ways is to apply data augmentations during training. 
However, recent work has shown that the set of augmentations usually used for terrestrial applications is not sufficient for orbital imagery~\cite{ulmer20236d}. 
To this end, we conduct an extensive study on the efficacy of different data augmentations to close the domain gap for spaceborne computer vision. 
We explore a range of augmentations, ranging from simple color augmentations, geometric transformations, image mixing, kernel filters, information deletion, corruptions, to noise.
In addition, we introduce two novel space augmentations that specifically aim to emulate visual conditions in orbit.
To investigate the importance of each augmentation, we utilize the hyperparameter optimization framework Optuna~\cite{optuna_2019}.
The framework automatizes the process of selecting different hyperparameter configurations for individual trials and allows to implement efficient search strategies.
Due to the sheer size of possible data augmentation combinations, we employ a \acrfull{TPE} algorithm to sample the search space.
As a reference task, we chose the field of 2D object detection as it is a well-studied and essential component of most spaceborne computer vision applications, such as pose estimation, inspection, and tracking. 
We train three very popular -- yet very different -- 2D object detectors: Mask R-CNN with a ResNet50 backbone, Faster R-CNN with a MobileNetv3 backbone, and \acrshort{yolo}-v7. 
We evaluate models and augmentations on the widely adopted SPEED+ dataset~\cite{Park2021-jr}. It's testing data is real recorded data from a real-world hardware-in-the-loop facility, hence, the focal point of the dataset is closing the domain gap.
In addition, we compare the performance to a recent open set object detector, called GroundingDINO~\cite{liu2023grounding}, that can be used out-of-the box and does not need to be fine-tuned or retrained. 
We evaluate all methods on an NVIDIA Jetson embedded system. 
Our findings demonstrate the importance of data augmentations to close the domain gap, leading to drastically improved performance and robustness to the challenging conditions of orbital imagery. 
At last, we present a set of augmentations that we found empirically to perform best to pave the way for more reliable and generalizable computer vision models for critical space applications. \\
To summarize, the contributions of this paper are: 
\begin{enumerate*}[label=(\roman*)]
    \item we demonstrate the severeness of the domain gap for object detection in orbit, 
    \item we conduct a large scale study involving 29 common augmentations and two novel space augmentations to find the best configuration for each detector,
    \item we explore the importance of the augmentations to each models and derive recommendations based on the collected data,
    \item we compare the performance to a modern open-set object detector and highlight differences in memory usage and inference speed on an NVIDIA Jetson embedded system.
\end{enumerate*}
\begin{figure*}[ht!]
    \centering
    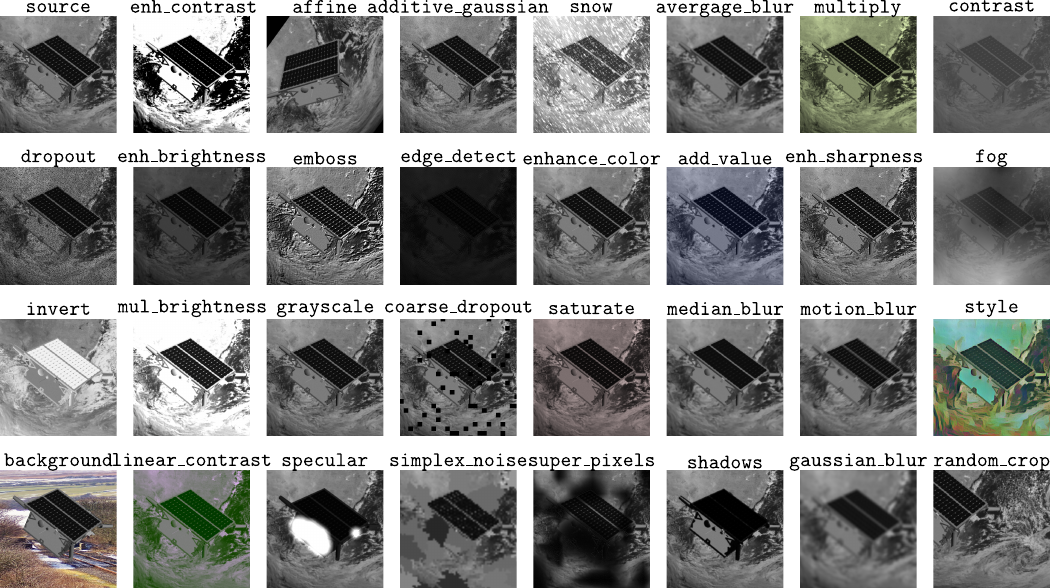
    \caption{Overview on all our possible data augmentations, applied to the same source image (leftmost image, in the first row).}
    \label{fig:augmentations}
\end{figure*}
\section{Related Work}
Methods based on deep neural networks have become widely adopted to solve computer vision tasks. One downside of such methods is their need of a large number of training samples to reach the desired performance~\cite{goodfellow2016deep}. 
The process of creating such datasets of real training data requires significant effort for terrestrial datasets~\cite{Hodan2017-gm} and is not practical for orbital perception. 
To this end, researchers started synthesizing images of 3D objects using computer graphics and train their models on such synthetic samples~\cite{Hinterstoisser2019-ah}. 
This approach allows the creation of massive datasets with only minimal human effort. Yet, neural networks trained on such synthetic datasets demonstrated a significant degradation in testing performance on real data~\cite{Rozantsev2019-nq}. 
This effect has been coined the domain gap and closing it is a central tenet of many modern computer vision methods~\cite{Sundermeyer2018-fl,Labbe2020-hb}. 

The SPEED and SPEED+ datasets showed that the domain gap is especially dramatic for space imagery~\cite{Park2021-jr}, likely due to the difficulty of rendering realistic orbital conditions. 
Many approaches were proposed to bridge this domain gap that can broadly be classified as domain adaption, domain randomization, and domain generalization.\\
\textit{Domain adaption} techniques aim to adapt trained models to the target domain at test time without annotations~\cite{Csurka2017-sd}. 
This means, that they have access to the target domain, yet, have no training labels. 
For instance, Wang \etal~\cite{Wang2024-ng} use images from the test set to train a \acrfull{GAN} to make training images more realistic. 
Park \etal~\cite{Park2023-ug} adapt the trained model to the target domain by unsupervised learning. The approach updates the parameters of normalization layers and affine transformations at test time by minimizing the shannon entropy of a segmentation head. 
At last, P{\'e}rez-Villar \etal~\cite{perez2022spacecraft} use a model trained on the source domain to generate pseudo annotation for the target domain. \\
\textit{Domain randomization} is another popular approach to bridging the domain gap by randomizing rendering parameters during training data synthesis. 
This approach has been shown to be very beneficial for object detection and pose estimation models in terrestrial applications~\cite{Tobin2017-qw,Denninger2020-cu}. 
However, to the best of our knowledge, such an approach has not been adopted by a published orbital dataset yet. \\
At last, \textit{domain generalization} is the process of setting up the training of the model that it generalizes to the target domain without any prior knowledge of the target domain~\cite{Wang2021-ou}. 
For instance, Li \etal~\cite{Li2018-gm} learn domain invariant latent features based on adverserial autoencoders. 
The most widely adopted implementation of domain generalization is data augmentation. 
The benefits can be manifold, such as increasing the training sample count, improving the training statistics variance, and reducing overfitting. 
Data augmentations have become a standard procedure in modern training of deep neural networks and most modern methods use it~\cite{Sundermeyer2018-fl,Labbe2020-hb,Hu2021-pl,ulmer20236d,wang2023yolov7,devries2017improved}.

Data augmentations for computer vision have been studied in the context of many different domains and tasks. 
In one of the earliest studies, Taylor \etal~\cite{Taylor2017-rx} study six classical data augmentations to improving performance on the Caltech101\footnote{https://data.caltech.edu/records/mzrjq-6wc02} image classification task. 
Mikołajczyk and Grochowski~\cite{Mikolajczyk2018-ku} focus on image classification of medical imagery and highlight a set of classical augmentations, texture transfer, and \acrshort{GAN}-based augmentations. 
Shorten \etal~\cite{Shorten2019-xy} present an extensive study on classical and deep-learning-based data augmentations, yet, is primarily focused on image classification. 
Hendrycks \etal~\cite{hendrycks2019benchmarking} benchmark neural network robustness to a list image corruption for image classification. 
In~\cite{Michaelis2019-hh} these corruptions are tested on the task of object detection for autonomous driving. 
Recently, Yang \etal~\cite{Yang2022-oi} evaluated classification, segmentation, and object detection methods on a fixed set of data augmentations and demonstrated that they improve inference performance. \\
Another interesting avenue of work to reduce the domain gap is to synthesize more realistic images. To this end, Hu \etal~\cite{Hu2021-pl} use a physics-based renderer with ray-tracing to simulate orbital conditions in the SwissCube dataset. \\
To the best of our knowledge, there has been no study focused on data augmentations for orbital imagery. Yet, Ulmer \etal~\cite{ulmer20236d} show that the augmentations that work best for terrestrial applications are not sufficient in the space domain. 
In this work, we do not attempt to find the best set of data augmentation manually but automate the process using a hyperparameter optimization framework~\cite{optuna_2019}. 
Moreover, we do not focus on a small set of augmentations, but optimize more than 30 different popular data augmentations. 
Due to this vast search space, we employ a search strategy that fits a \acrfull{GMM} to the experiment results and conduct hundreds of experiments to explore the importance of data augmentations for orbital perception.

\section{Method}
In this section, we first outline the task of object detection and the unique characteristics of the detectors that we deploy.
Next, we discuss the set of data augmentation techniques which we investigate in the context of satellite object detection.
Eventually, and for the means of our data augmentation investigation, we introduce the hyperparameter optimization strategy to find the best augmentation set.
The complete optimization framework is shown in \cref{fig:overview} and can be described as follows.
The process runs for $n_{\text{trials}}$ trials and a sampler chooses a set of hyperparameters $h_i$ with $i \in \bigl[0, n_{\text{trials}}-1 \bigr]$. Each set $h_i$ parameterizes a chain of augmentations that is applied to each image $\boldsymbol{I}$ during training $\boldsymbol{\hat{I}}=\texttt{augmenter}_{h_i} \bigl(\boldsymbol{I} \bigr)$. The parameterization of an \texttt{augmenter} remains fixed for one full trial. In each trial, we train a $\texttt{detector}_{\theta_i}$ that is parameterized by its weights $\theta_i$. After a fixed amount of training epochs is complete, we evaluate the $\texttt{detector}_{\theta_i}$ on the test dataset and calculate an evaluation score $\acrshort{mAP}_i$. Then, we store the pair $(h_i, \acrshort{mAP}_i)$ in a database. For the initial $n_{\text{startup\_trials}}$ the sampler randomly chooses $h_i$ to initialize the \acrshort{GMM}. After the startup phase is complete, the sampler uses the database and the \acrshort{TPE} search strategy to generate predicted high importance parameters $h_i$.

\subsection{2D Object Detection}
Given an image, $\boldsymbol{I} \in \mathbb{R}^{\text{w} \times \text{h} \times \text{3}}$ with width $\text{w}$ and height $\text{h}$, the goal of 2D object detection is to recognize all instances of an object class and localize them by their axis-aligned bounding boxes~\cite{Zaidi2021-ab}. 
In this work, we investigate the performance of three different object detection groups in orbital visual conditions. 
First, the family of region-based detectors which predict objects and their bounding boxes by processing images in two distinct stages~\cite{Ren2015-ly}. 
Second, and in contrast to the former category, single-stage detectors directly regress objects and their locations. 
These two approaches detect objects from a pre-defined object set on which they are specifically trained on.
This differs to the third group, open-set detectors, which require no additional training on specific objects and thus can be deployed out-of-the-box. 

\subsubsection{Region-based Object Detection}
These detectors consist of two stages: first a region-proposal stage which outputs regions of interest for the second stage, a regression model that estimates the actual object class and bounding box coordinates in the input image. 
In~\cite{Ren2015-ly}, Ren \etal~introduce a learnable neural network for the region-proposal stage, predicting object scores and anchor boxes with varying dimensions to account for different object sizes and scales. 
Based on these regions of interest, the consecutive detection model learns to regress the actual object class labels and bounding boxes.
In this work, we train two region-based detectors: Faster R-CNN~\cite{Ren2015-ly} with a MobileNet~\cite{Howard2017MobileNetsEC} backbone and Mask R-CNN~\cite{8237584} with a ResNet50-based~\cite{He2015DeepRL} feature extractor.
The latter extends Faster R-CNN by simultaneously predicting segmentation masks for detected objects through expanding the architecture with an additional segmentation head.
We train these two detectors with different backbones as this affects the final network size, making Faster R-CNN smaller than Mask R-CNN.
~\cref{tab:training_parameters} shows the exact model sizes as presented by the number of parameters.
The rationale is that smaller sized networks require less compute and memory -- a consideration relevant for the space domain. 
However, as we discuss in~\cref{sec:exp}, the network size and parameter quantity influence its capacity to learn patterns and thus can impact performance. 

\subsubsection{Single-Stage Object Detection}
This group of object detection networks mainly consists of \acrfull{yolo}~\cite{Redmon2015YouOL} models and its variants and extensions \cite{Xu2022PPYOLOEAE,Ge2021YOLOXEY,wang2023you,wang2023yolov7}.
Single-stage detection refers to the simultaneous and direct regression of bounding box coordinates and class probabilities from a given input image.
Therefore, the image $\boldsymbol{I}$ is split into a grid of $s \times s$ cells and each cell predicts whether an object center falls into its cell~\cite{Redmon2015YouOL}.
Casting object detection as regression problem enables bounding boxes predictions at a high inference speed compared to region-based approaches. 
For our experiments, we train a \acrshort{yolo}-v7~\cite{wang2023yolov7} network with almost $8$ times more trainable parameters (see~\cref{tab:training_parameters}) than the Faster R-CNN -- and hence regarded as more powerful. 
This advanced \acrshort{yolo} architecture introduces model re-parameterization and model scaling, which improves the detection accuracy while maintaining its inference speed~\cite{wang2023yolov7}. 
Additionally, by adjusting the training routine and adding an auxiliary loss, Wang \etal~further increase the detection accuracy.

\subsubsection{Open-Set Object Detection} \label{subsubsec:grounding}
Detectors of this category can identify unknown objects, \ie unseen during the model's training phase, in addition to known, trained on objects.
Therefore, recent models take as an additional input a descriptive text prompt about the objects that are supposed to be detected in an image~\cite{guopen,Kuo2022FindItGL,liu2023grounding}.
Consequently, aligning visual and language modalities becomes necessary.
In this work, we apply GroundingDINO~\cite{liu2023grounding}, which is open-source and achieves competitive results on various detection benchmarks. 
GroundingDINO approaches the modality alignment through a large-scale pre-training phase on known datasets and objects.
Furthermore, Liu \etal introduce an approach to improve the mapping of text features to specific image regions for a better modality fusion.
In contrast to region-based and single-stage approaches, open-set detectors do not require further training, which poses an advantage.
However, since it requires a text prompt as an additional input, the prompt selection directly impacts the detection performance -- an observation we report and further elaborate in~\cref{sec:exp}.

\subsection{Data Augmentations}
The application of data augmentations to the training dataset has long been a standard practice to train neural networks. 
Such models are very prone to overfitting and require proper regularization to generalize beyond the training data~\cite{devries2017improved}.
Besides regularization, another goal of augmenting the training data is to increase the variance, improving alignment between the inference and training distributions. 
In the context of orbital perception, simulating visual conditions such as specular reflections is very difficult and costly. 
Hence, synthetic images hardly entail realistic lighting conditions and models perform poorly on such samples which are clearly beyond the training distribution~\cite{ulmer20236d}.
In this work, we study the impact of five commonly used classes~\cite{Yang2022-oi} of image data augmentations: color, geometric, mixing, filters, and deletion.
We further introduce specialized augmentations tailored to the distinctive characteristics of the space environment.

\subsubsection{Color augmentations} change the statistics of the color channels of an image. 
\texttt{add\_value} selects a random value in a set interval and adds this value to all or individual channels. 
\texttt{invert} inverts pixel values for all or individual channels.
\texttt{multiply} selects a random value in a set interval and multiplies this value to all or individual channels. 
\texttt{multiply\_brightness} convert an image to the HSV colorspace and multiplies a random value in a set interval to the brightness-related channel before converting the image back to the original colorspace. \texttt{enhance\_color} changes the balance of the color channels of an image. 
\texttt{grayscale} converts the image to grayscale and overlays the gray version with the original by a random strength in a set interval. 
\texttt{contrast} changes the contrast of the image by a set severity~\cite{hendrycks2019benchmarking}. 
\texttt{linear\_contrast} modifies the contrast of an image linearly with a randomly chosen value in a set interval. 
\texttt{enhance\_contrast} increases or decreases the contrast of an image randomly in a set interval. \texttt{saturate} changes the saturation of an image, making it randomly more or less colorful~\cite{hendrycks2019benchmarking}.
At last, \texttt{enhance\_brightness} increases or decreases the brightness of an image randomly in a set interval.

\subsubsection{Geometric augmentations} apply some geometric transformation to the image, fundamentally changing it.
\texttt{affine} applies an affine transformation to the image, such as scaling, translation, rotation, and shearing. In this work, we only apply a random rotation in a set interval. 
\texttt{random\_crop} calculates a random scale change and crops the image randomly such that a minimum area of the target object is still visible. 

\subsubsection{Image mixing} changes the contents of an image by mixing it with some other source. 
\texttt{background} uses the binary foreground segmentation to replace the background of the image with some randomly sampled image, often from another dataset~\cite{Sundermeyer2018-fl,Everingham2010-cu} .
\texttt{snow} adds randomly generated snow layers with a set severity from random directions which obstruct visual features~\cite{hendrycks2019benchmarking}.
\texttt{fog} adds a randomly generate layer of fog to the image with a set severity.

\subsubsection{Kernel filters} apply some transformation to the image by applying a kernel or convolution operation. 
\texttt{gaussian\_blur} applies a gaussian kernel with a standard deviation randomly chosen in a set interval.
\texttt{average\_blur} computes means over a neighbourhood randomly chosen in a set interval.
\texttt{median\_blur} computes the median over a neighbourhood randomly chosen in a set interval.
\texttt{motion\_blur} convolves the image with a random kernel that simulates motion blur of a moving camera or object. 
\texttt{emboss} calculates an embossed version of the image that pronounces highlights and shadows and blends the result randomly with the original.
\texttt{edge\_detect} convolves the original with an edge-detection kernel and then randomly blends the result with the original.
\texttt{enhance\_sharpness} randomly increases or decreases the sharpness of an image.
\texttt{additiv\_guassian\_noise} adds noise that is sampled from a normal gaussian distribution with random variance elementwise to the image.
\texttt{style} augmentation is a neural network which is trained to transform an image such that it remains semantically valid but randomizes the style of the image~\cite{Jackson2018-ye}.

\subsubsection{Information deletion} erases data from the image by some function. 
\texttt{super\_pixels} chooses a random number of super pixels per image and replaces the value with a randomly chosen probability by its average pixel color.
\texttt{simplex\_noise} creates a simplex noise mask with that is used to blend the original image with the result of an edge-detection operation.
\texttt{dropout} sets individual pixel values at random locations to zero for a randomly sampled fraction of the image, for all or individual channels.
\texttt{coarse\_dropout} chooses rectangular areas to drop at random locations for a randomly sampled fraction of the image, for all or individual channels.

\subsubsection{Space augmentations} 
\begin{figure}
    \centering
    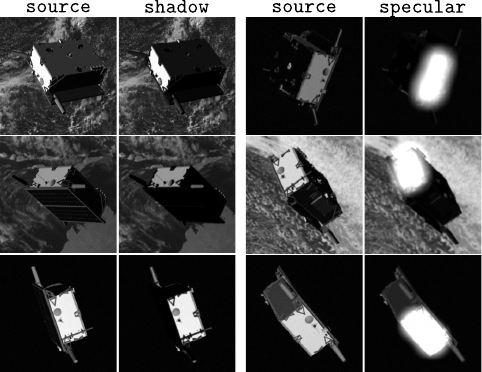
    \caption{Visualization of our custom space data augmentations, simulating typical visual conditions encountered on objects in orbit: stark \texttt{shadow} effects and strong \texttt{specular} reflections. 
    Source images displayed in the left, augmented ones in the right columns.}
    \label{fig:aug_space}
\end{figure}

attempt to emulate the special visual conditions of orbital perception. 
These augmentations are developed by Ulmer \etal\cite{ulmer20236d} in the scope of the SPEED+ challenge to realize space specific data augmentations. \texttt{specular} simulates reflections which can occur on metallic surfaces and result in bloom-like visual artifacts that visually occlude features. 
Such effects are usually confined locally to the target satellite and the rest of the image is largely unchanged.
To this end, the method uses the binary segmentation mask to sample the brightest region of the foreground and overlays a white bloom, bleeding into the rest of the image. 
\texttt{shadow} emulates the exceedance of the dynamic range of a camera by selecting the dark regions of the target, making the even darker while keeping the bright region constant.
Examples of these augmentations are visualized in~\cref{fig:aug_space}.

\subsection{Hyperparameter Optimization}
Our goal is to find a good set of hyperparameters in as little trials as possible. Finding the best set of hyperparameters in such a large space is very difficult due to time and resource constraints. Even if each parameter is only considered to be binary -- either active or inactive -- the search space quickly reaches millions of possible trials which makes an approach like grid search impossible. 
Historically, such an optimization has been conducted by humans in-the-loop and the effectiveness and results comes down to the experience of the researcher tweaking a few parameters at a time. However, modern software and hardware make it possible to automate the process and run many trials in parallel on compute clusters. 

In this work, we use the Optuna framework~\cite{optuna_2019} to conduct the exploration of the data augmentations search space. Modelling the dynamics of data augmentations is an inherently difficult task for any optimization process. As we will show, too little data augmentations result in poor performance to close the domain gap. Yet, too many augmentations can have detrimental effects on the performance due to degradation of information, often to the point that no training signal is retained in individual samples. This can lead to highly unstable gradients and catastrophic training dynamics. In addition, some augmentations have a strong relational component where combinations have a reinforcing or deteriorating effect.

To this end, we use the \acrfull{TPE}~\cite{Bergstra2011-jx} to sample the search space. \acrshort{TPE} is a Bayesian optimization method that can handle tree-structured search spaces and uses a kernel density estimator. In short, the estimator fits one \acrfull{GMM} to the parameters correlated with the best objective results, and another \acrshort{GMM} to the remaining hyperparameters. For more information, please refer to \cite{Watanabe2023-xz}. The estimator is initialized for $n_\text{startup\_trials}$ random trials for which the objective values are stored in a database. After the startup phase, the sampler draws a new hyperparameter configurations for each new trial based on its internal model.

\section{Experiment}\label{sec:exp}
\begin{figure}[htb!]
    \includegraphics[width=0.48\textwidth]{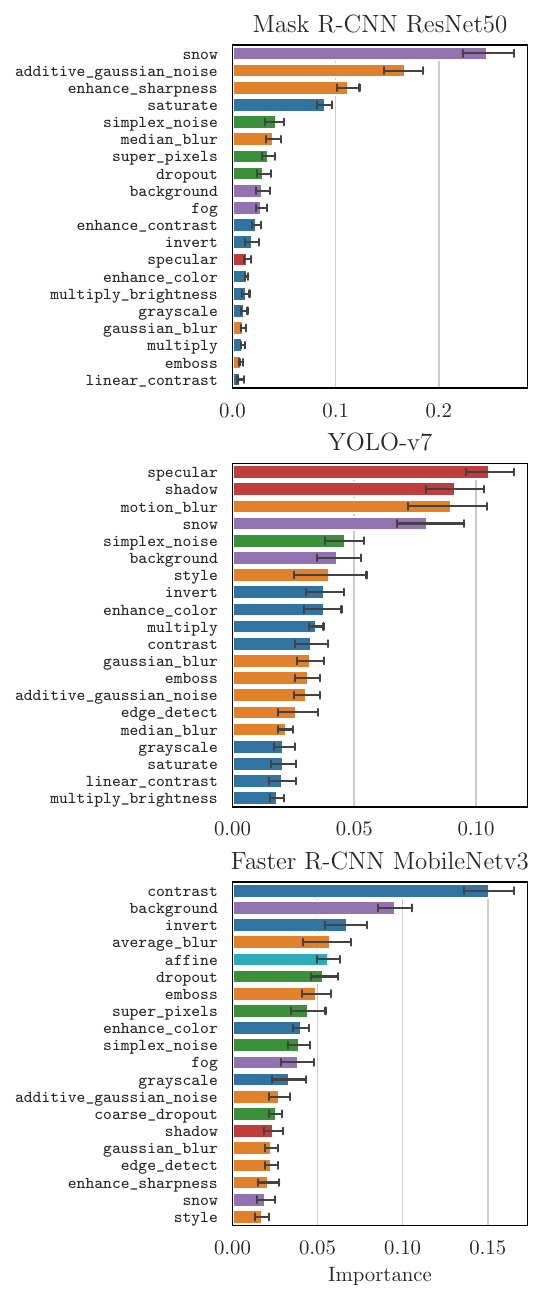}
    \caption{Importance and variance of individual data augmentations on the different models' precision performance.
    We observe varying key augmentations for each detection model. 
    Furthermore, network size seems to impact whether severe image changing techniques benefit a model, with the larger \acrshort{yolo}-v7 being the only detector for which those augmentations are more influential.}
    \label{fig:importance}
\end{figure}
In this section, we first describe the dataset that is the basis for our results and the changes we employed to the original data. 
Next, we cover the setup of the training pipeline for the models and the hyperparameter optimization framework, before exploring the results of this study.
\begin{table*}[htb!]
\renewcommand{\arraystretch}{1.3}
\caption{Training parameters of all models that were fixed for all training runs.}
\begin{center}
\begin{tabular}{|lccc|}
\hline
& Mask R-CNN ResNet50 & YOLO-v7 & Faster R-CNN MobileNetv3  \\
\hline
epochs & 20 & 20 & 20 \\
avg. train walltime [h]  & 15.08 & 18.94 & 10.88 \\
\#parameters [M] & 43.7 & 164.89 & 18.87 \\
batch size & 16 & 16 & 16 \\
optimizer & SGD & SGD &  SGD \\
learning rate & 0.02 & 0.01 & 0.02  \\
momentum &  0.9 & 0.937 & 0.9  \\
weight decay & \num{1e-4} & \num{1e-5} & \num{1e-4}  \\
warmup steps & 2000 & 2000 & 2000  \\
annealing point & 0.72 & 0.72 & 0.72 \\
\hline
\end{tabular}
\label{tab:training_parameters}
\end{center}
\end{table*}

\subsection{SPEED+ Satellite Pose Estimation Challenge}
The Next-Generation \underline{S}pacecraft \underline{P}os\underline{E} \underline{E}stimation \underline{D}ataset (SPEED+) was the subject of the 6D pose estimation challenge \gls{SPEC} organized by the \gls{ESA} and Stanford University~\cite{Park2021-jr}. 
The purpose of the competition was to determine the current state-of-the-art performance on the challenging task of satellite pose estimation and closing the domain gap. 
The training and validation split of the dataset are $59960$ computer-rendered images of the Tango satellite annotated with the 6D pose.
For testing, the dataset contains $9531$ \gls{HIL} mockup-satellite images split into two subsets 
with distinct visual conditions.
Concretely, the \texttt{lightbox} split simulates albedo conditions using light boxes with diffuser plates and the \texttt{sunlamp} split entails a high-intensity sun simulator to mimic direct homogeneous light from the sun~\cite{Park2021-jr}.

In this work, we repurpose the dataset to 2D object detection. The original data does not include a 3D model of the target nor 2D detection or segmentation annotations for training and test set. 
To this end, we established a coarse wireframe model of the satellite by inspection and used this faulty model in combination with the given 6D pose to generate the missing annotations for the original training images, similar to \cite{ulmer20236d}. 
In addition, we rectified each sample and reduce the resolution to $960 \times 600$ pixels.

\subsection{Experiment Setup}
We trained three very different object detectors for this study. 
Modern methods often employ a variety of techniques to improve learning performance and stability, such as \gls{EMA} and elaborate learning rate schedules, that are fine-tuned to the specific model. 
To keep the results as comparable as possible we settled on a simple training schedule of $20$ epochs with a linear learning rate warmup and cosine annealing to finish the training. 
We kept the batch size, learning rate and optimizer configuration  as close as possible to the default values of the original works and references. 
A full list of all training parameters can be found in~\cref{tab:training_parameters}. 
All models were trained on a single NVIDIA A100 GPU with 80 GB of VRAM, however, the Mask R-CNN and Faster R-CNN would have fit on a smaller GPU.
\begin{table*}
\caption{Our satellite detection evaluation on SPEED+ images with varying data augmentations applied during training.}
\renewcommand{\arraystretch}{1.3}
\centering
\addtolength{\tabcolsep}{-0.4em}
\begin{tabular}{lcccccccccccccccccccccccccccccccccccc} 
$\sum$ & \rot{\texttt{affine}} & \rot{\texttt{shadow}} & \rot{\texttt{specular}} & \rot{\texttt{background}} & \rot{\texttt{random\_crop}} & \rot{\texttt{fog}} & \rot{\texttt{snow}} & \rot{\texttt{emboss}} & \rot{\texttt{invert}} & \rot{\texttt{dropout}} & \rot{\texttt{contrast}} & \rot{\texttt{multiply}} & \rot{\texttt{saturate}} & \rot{\texttt{add\_value}} & \rot{\texttt{grayscale}} & \rot{\texttt{edge\_detect}} & \rot{\texttt{median\_blur}} & \rot{\texttt{motion\_blur}} & \rot{\texttt{average\_blur}} & \rot{\texttt{super\_pixels}} & \rot{\texttt{enhance\_color}} & \rot{\texttt{gaussian\_blur}} & \rot{\texttt{simplex\_noise}} & \rot{\texttt{coarse\_dropout}} & \rot{\texttt{linear\_contrast}} & \rot{\texttt{enhance\_contrast}} & \rot{\texttt{enhance\_sharpness}} & \rot{\texttt{enhance\_brightness}} & \rot{\texttt{multiply\_brightness}} & \rot{\texttt{additive\_gaussian}} & \rot{\texttt{style}} & \rott{$\text{mAP}_{lightbox}$} & \rott{$\text{mAP}_{sunlamp}$} & \rott{$\text{IoU}$} &  \rott{$\text{mAP}@75$} & \rott{$\text{mAP}$} \\ 
\hline
\multicolumn{37}{|c|}{Mask R-CNN ResNet50} \\
\hline 
18 & \checkmark & \checkmark & 0 & 0 & \checkmark & 0 & \checkmark & \checkmark & 0 & 0 & \checkmark & \checkmark & \checkmark & 0 & \checkmark & 0 & \checkmark & \checkmark & 0 & 0 & \checkmark & 0 & \checkmark & \checkmark & \checkmark & 0 & \checkmark & \checkmark & 0 & \checkmark & 0 & 0.881 & 0.827 & 0.918 & 0.965 & 0.867 \\
18 & \checkmark & \checkmark & \checkmark & 0 & \checkmark & \checkmark & \checkmark & 0 & \checkmark & \checkmark & \checkmark & \checkmark & \checkmark & 0 & \checkmark & 0 & \checkmark & 0 & \checkmark & 0 & 0 & 0 & 0 & 0 & 0 & 0 & \checkmark & 0 & \checkmark & \checkmark & \checkmark & 0.885 & 0.829 & 0.920 & 0.968 & 0.869 \\
15 & 0 & \checkmark & 0 & \checkmark & \checkmark & \checkmark & \checkmark & 0 & \checkmark & \checkmark & 0 & 0 & 0 & 0 & \checkmark & 0 & \checkmark & \checkmark & \checkmark & 0 & 0 & 0 & \checkmark & 0 & \checkmark & 0 & 0 & \checkmark & 0 & \checkmark & 0 & 0.884 & 0.832 & 0.918 & 0.968 & 0.869 \\
20 & \checkmark & \checkmark & \checkmark & 0 & \checkmark & \checkmark & \checkmark & \checkmark & \checkmark & \checkmark & 0 & \checkmark & \checkmark & 0 & 0 & 0 & \checkmark & 0 & 0 & 0 & \checkmark & \checkmark & 0 & 0 & \checkmark & \checkmark & \checkmark & \checkmark & \checkmark & \checkmark & 0 & 0.888 & 0.823 & 0.920 & 0.968 & 0.869 \\
20 & \checkmark & \checkmark & \checkmark & 0 & \checkmark & \checkmark & \checkmark & 0 & \checkmark & \checkmark & 0 & \checkmark & \checkmark & 0 & \checkmark & 0 & \checkmark & \checkmark & 0 & 0 & \checkmark & 0 & \checkmark & 0 & \checkmark & 0 & \checkmark & \checkmark & \checkmark & \checkmark & 0 & 0.886 & 0.841 & 0.922 & 0.967 & 0.873 \\
19 & \checkmark & \checkmark & \checkmark & 0 & \checkmark & \checkmark & \checkmark & 0 & \checkmark & \checkmark & 0 & 0 & \checkmark & 0 & \checkmark & \checkmark & 0 & \checkmark & 0 & 0 & \checkmark & 0 & \checkmark & 0 & \checkmark & 0 & \checkmark & \checkmark & \checkmark & \checkmark & 0 & 0.891 & 0.842 & \textbf{0.925} & 0.969 & 0.876 \\
21 & \checkmark & \checkmark & \checkmark & 0 & \checkmark & \checkmark & \checkmark & 0 & \checkmark & \checkmark & 0 & \checkmark & \checkmark & 0 & \checkmark & \checkmark & 0 & \checkmark & 0 & 0 & \checkmark & 0 & \checkmark & 0 & \checkmark & \checkmark & \checkmark & \checkmark & \checkmark & \checkmark & 0 & 0.891 & \textbf{0.845} & \textbf{0.925} & 0.968 & 0.877 \\
20 & \checkmark & \checkmark & \checkmark & 0 & \checkmark & \checkmark & \checkmark & 0 & \checkmark & \checkmark & 0 & \checkmark & \checkmark & 0 & \checkmark & \checkmark & 0 & \checkmark & 0 & 0 & \checkmark & 0 & \checkmark & 0 & \checkmark & 0 & \checkmark & \checkmark & \checkmark & \checkmark & 0 & 0.892 & 0.843 & \textbf{0.925} & 0.971 & 0.877 \\
21 & \checkmark & \checkmark & \checkmark & 0 & \checkmark & \checkmark & \checkmark & 0 & \checkmark & \checkmark & 0 & \checkmark & \checkmark & 0 & \checkmark & \checkmark & \checkmark & \checkmark & 0 & 0 & \checkmark & 0 & \checkmark & 0 & \checkmark & 0 & \checkmark & \checkmark & \checkmark & \checkmark & 0 & 0.893 & 0.844 & \textbf{0.925} & 0.973 & 0.878 \\
18 & \checkmark & \checkmark & \checkmark & \checkmark & \checkmark & \checkmark & \checkmark & 0 & \checkmark & \checkmark & 0 & 0 & \checkmark & 0 & \checkmark & \checkmark & 0 & \checkmark & 0 & 0 & \checkmark & 0 & \checkmark & 0 & 0 & \checkmark & \checkmark & 0 & 0 & \checkmark & 0 & \textbf{0.895} & 0.843 & \textbf{0.925} & \textbf{0.977} & \textbf{0.881} \\
\hline 
\hline 
\multicolumn{37}{|c|}{YOLO-v7} \\
\hline
14 & 0 & 0 & 0 & \checkmark & \checkmark & 0 & \checkmark & \checkmark & 0 & \checkmark & 0 & \checkmark & 0 & 0 & \checkmark & 0 & \checkmark & 0 & 0 & \checkmark & \checkmark & 0 & \checkmark & 0 & \checkmark & \checkmark & \checkmark & 0 & 0 & 0 & 0 & 0.918 & 0.873 & 0.935 & 0.957 & 0.904 \\
16 & \checkmark & \checkmark & \checkmark & \checkmark & \checkmark & 0 & \checkmark & 0 & 0 & \checkmark & \checkmark & 0 & 0 & 0 & 0 & \checkmark & \checkmark & 0 & \checkmark & \checkmark & \checkmark & \checkmark & 0 & 0 & 0 & \checkmark & 0 & 0 & \checkmark & 0 & 0 & 0.926 & 0.850 & 0.934 & 0.958 & 0.904 \\
17 & 0 & \checkmark & 0 & \checkmark & \checkmark & \checkmark & \checkmark & 0 & 0 & \checkmark & \checkmark & 0 & 0 & 0 & \checkmark & 0 & \checkmark & 0 & 0 & 0 & \checkmark & 0 & \checkmark & 0 & \checkmark & \checkmark & \checkmark & \checkmark & 0 & \checkmark & \checkmark & 0.917 & 0.888 & 0.937 & 0.968 & 0.907 \\
10 & \checkmark & 0 & 0 & \checkmark & \checkmark & 0 & \checkmark & 0 & \checkmark & 0 & \checkmark & 0 & 0 & 0 & 0 & 0 & 0 & \checkmark & 0 & 0 & \checkmark & 0 & 0 & 0 & 0 & 0 & 0 & \checkmark & 0 & \checkmark & 0 & 0.927 & 0.876 & 0.938 & 0.959 & 0.912 \\
16 & 0 & 0 & \checkmark & \checkmark & \checkmark & 0 & 0 & \checkmark & 0 & 0 & \checkmark & \checkmark & 0 & 0 & \checkmark & \checkmark & \checkmark & 0 & 0 & \checkmark & \checkmark & \checkmark & 0 & 0 & \checkmark & \checkmark & 0 & 0 & 0 & \checkmark & \checkmark & 0.920 & 0.895 & 0.937 & 0.969 & 0.912 \\
19 & 0 & \checkmark & 0 & \checkmark & \checkmark & \checkmark & \checkmark & 0 & \checkmark & \checkmark & 0 & 0 & 0 & 0 & \checkmark & \checkmark & \checkmark & 0 & \checkmark & \checkmark & \checkmark & 0 & \checkmark & 0 & \checkmark & \checkmark & \checkmark & \checkmark & 0 & 0 & \checkmark & 0.924 & 0.890 & 0.940 & 0.968 & 0.914 \\
18 & 0 & \checkmark & 0 & \checkmark & \checkmark & \checkmark & \checkmark & \checkmark & 0 & \checkmark & \checkmark & 0 & 0 & 0 & 0 & 0 & \checkmark & \checkmark & 0 & \checkmark & 0 & \checkmark & \checkmark & \checkmark & \checkmark & \checkmark & \checkmark & 0 & 0 & \checkmark & 0 & 0.941 & 0.858 & 0.940 & 0.969 & 0.918 \\
17 & 0 & \checkmark & \checkmark & \checkmark & \checkmark & \checkmark & \checkmark & 0 & 0 & \checkmark & \checkmark & \checkmark & \checkmark & 0 & 0 & \checkmark & \checkmark & 0 & 0 & 0 & \checkmark & 0 & 0 & 0 & \checkmark & \checkmark & 0 & 0 & \checkmark & 0 & \checkmark & 0.932 & 0.907 & 0.945 & \textbf{0.979} & 0.925 \\
15 & \checkmark & 0 & 0 & \checkmark & \checkmark & \checkmark & \checkmark & 0 & 0 & 0 & \checkmark & 0 & 0 & 0 & 0 & \checkmark & 0 & \checkmark & 0 & 0 & 0 & \checkmark & \checkmark & 0 & \checkmark & \checkmark & 0 & \checkmark & 0 & \checkmark & \checkmark & \textbf{0.942} & 0.900 & 0.947 & \textbf{0.979} & 0.931 \\
16 & \checkmark & 0 & \checkmark & \checkmark & \checkmark & \checkmark & \checkmark & 0 & 0 & 0 & \checkmark & \checkmark & \checkmark & 0 & \checkmark & 0 & 0 & 0 & 0 & 0 & 0 & 0 & \checkmark & 0 & \checkmark & 0 & \checkmark & \checkmark & 0 & \checkmark & \checkmark & 0.937 & \textbf{0.919} & \textbf{0.948} & \textbf{0.979} & \textbf{0.932} \\
\hline 
\hline 
\multicolumn{37}{|c|}{Faster R-CNN MobileNetv3} \\
\hline
17 & \checkmark & \checkmark & \checkmark & \checkmark & \checkmark & 0 & 0 & \checkmark & \checkmark & \checkmark & \checkmark & 0 & 0 & 0 & 0 & \checkmark & \checkmark & 0 & \checkmark & 0 & 0 & \checkmark & \checkmark & \checkmark & 0 & 0 & \checkmark & 0 & 0 & 0 & \checkmark & 0.726 & 0.730 & 0.858 & 0.850 & 0.726 \\
13 & 0 & 0 & \checkmark & \checkmark & \checkmark & \checkmark & 0 & \checkmark & 0 & \checkmark & \checkmark & \checkmark & \checkmark & 0 & 0 & 0 & 0 & 0 & 0 & \checkmark & 0 & 0 & 0 & 0 & \checkmark & 0 & \checkmark & 0 & 0 & \checkmark & 0 & 0.737 & 0.709 & 0.846 & 0.856 & 0.728 \\
17 & \checkmark & 0 & \checkmark & \checkmark & \checkmark & 0 & 0 & 0 & \checkmark & \checkmark & \checkmark & \checkmark & 0 & 0 & \checkmark & \checkmark & \checkmark & 0 & \checkmark & 0 & 0 & 0 & \checkmark & \checkmark & 0 & \checkmark & \checkmark & 0 & 0 & \checkmark & 0 & 0.732 & 0.723 & 0.860 & 0.858 & 0.729 \\
18 & 0 & 0 & \checkmark & 0 & \checkmark & \checkmark & 0 & \checkmark & \checkmark & \checkmark & \checkmark & \checkmark & 0 & 0 & 0 & \checkmark & \checkmark & 0 & \checkmark & 0 & \checkmark & 0 & 0 & 0 & \checkmark & 0 & \checkmark & \checkmark & \checkmark & \checkmark & \checkmark & 0.740 & 0.714 & 0.852 & 0.846 & 0.731 \\
18 & \checkmark & \checkmark & 0 & \checkmark & \checkmark & \checkmark & \checkmark & \checkmark & 0 & \checkmark & \checkmark & \checkmark & 0 & 0 & \checkmark & \checkmark & 0 & \checkmark & \checkmark & 0 & 0 & 0 & \checkmark & 0 & 0 & \checkmark & \checkmark & 0 & 0 & 0 & \checkmark & 0.741 & 0.713 & 0.858 & 0.856 & 0.732 \\
18 & \checkmark & 0 & \checkmark & \checkmark & \checkmark & 0 & \checkmark & 0 & 0 & \checkmark & \checkmark & 0 & \checkmark & 0 & \checkmark & \checkmark & \checkmark & 0 & 0 & 0 & \checkmark & 0 & \checkmark & \checkmark & \checkmark & 0 & \checkmark & 0 & \checkmark & 0 & \checkmark & 0.736 & \textbf{0.740} & 0.861 & 0.862 & 0.737 \\
15 & 0 & \checkmark & \checkmark & 0 & \checkmark & 0 & \checkmark & 0 & \checkmark & \checkmark & 0 & 0 & 0 & 0 & 0 & 0 & 0 & \checkmark & \checkmark & \checkmark & 0 & \checkmark & \checkmark & 0 & \checkmark & \checkmark & 0 & \checkmark & 0 & \checkmark & 0 & 0.748 & 0.714 & 0.860 & 0.869 & 0.739 \\
18 & \checkmark & \checkmark & \checkmark & 0 & \checkmark & 0 & \checkmark & \checkmark & \checkmark & 0 & \checkmark & \checkmark & 0 & 0 & \checkmark & 0 & \checkmark & 0 & \checkmark & 0 & \checkmark & 0 & \checkmark & \checkmark & \checkmark & \checkmark & 0 & 0 & 0 & \checkmark & 0 & 0.743 & 0.734 & 0.860 & 0.871 & 0.740 \\
19 & \checkmark & \checkmark & 0 & \checkmark & \checkmark & 0 & 0 & 0 & \checkmark & \checkmark & \checkmark & 0 & \checkmark & 0 & 0 & 0 & \checkmark & \checkmark & 0 & \checkmark & \checkmark & \checkmark & \checkmark & 0 & \checkmark & 0 & \checkmark & \checkmark & \checkmark & \checkmark & 0 & 0.753 & 0.720 & 0.862 & 0.872 & 0.743 \\
19 & \checkmark & \checkmark & 0 & \checkmark & \checkmark & \checkmark & \checkmark & \checkmark & \checkmark & \checkmark & \checkmark & 0 & \checkmark & 0 & 0 & 0 & 0 & \checkmark & \checkmark & 0 & 0 & 0 & \checkmark & 0 & \checkmark & \checkmark & 0 & 0 & \checkmark & \checkmark & \checkmark & \textbf{0.755} & 0.739 & \textbf{0.865} & \textbf{0.883} & \textbf{0.750} \\
\hline
\hline
\end{tabular}
\label{tab:overall_performance}
\end{table*}

For all three models we set up Optuna to conduct the hyperparameter optimization in an identical fashion. 
We initialize a database for each model and run $n_\text{startup\_trials}=64$ trials to initialize the \acrshort{TPE}. 
To keep the search space as small as possible, each augmentation probability is defined by a categorical distribution to be either active or inactive. 
If an augmentation is chosen to be active in a trial, we apply the augmentation to an image with a probability of $30\%$. 
For all data augmentations,we defined a default parameterization that is used whenever the augmentation is active that can be found in the github repository. 
For the objective metric we use the \acrfull{mAP} calculated over the steps $[0.5,...,0.95]$ with step size $0.05$.

\subsection{Results}
In this subsection we first explore each augmentation factor using the \acrfull{fANOVA} framework to quantify the importance of individual hyperparameters, before presenting the best set of hyperparameters that we found for each individual model.
\subsubsection{Hyperparameter Importance} First, we try to gain some insights into the relative importance of different data augmentations for each model. The \acrshort{fANOVA} algorithm fits a random forest to the hyperparameter space that regresses the objective value, in our case the \acrshort{mAP}. We run the algorithm with $64$ trees in the forest and a maximum depth of $64$. In addition, we repeat the algorithm $8$ times to assess the variance in the results from the algorithm. The $20$ most important augmentations, summing up to one, for each model are shown in~\cref{fig:importance}.

For Mask R-CNN, the first four factors explain about 60\% of the variance in the data we collected. These augmentations are \texttt{snow}, \texttt{additive\_gaussian\_noise}, \texttt{enhance\_sharpness}, and \texttt{saturate}. 
The variance between different runs of \acrshort{fANOVA} is low. 
This suggest that the random forest model can represent the data well and the \acrshort{TPE} has converged to some local maximum. 
There is a second tier of augmentations that explain another 25\% of the variance, containing \texttt{dropout}, \texttt{fog}, \texttt{enhance\_contrast}, and \texttt{invert} among others. 
The augmentations that scored the worst in terms of importance, all below 0.5\%, are \texttt{average\_blur}, \texttt{motion\_blur}, \texttt{style}, \texttt{enhance\_brightness}, \texttt{coarse\_dropout}, and \texttt{affine}.

Analysing \acrshort{yolo}-v7, the results of the \acrshort{fANOVA} algorithm are more uncertain. The results of different random forests vary more than they did for Mask R-CNN. However, there are four factors which clearly express more variance than the rest, about 36\%.
Namely, \texttt{specular}, \texttt{shadow}, \texttt{motion\_blur}, and \texttt{snow}. 
Notably, the \texttt{style} augmentation~\cite{Jackson2018-ye} varies extremely between different runs. 
We hypothesize, that \texttt{style}, which is a pretrained deep convolution network, has strong interactions with other data augmentations. 
In our experiments, the least important data augmentations for \acrshort{yolo}-v7 are \texttt{fog}, \texttt{average\_blur}, and \texttt{enhance\_contrast}, all with around 1\% importance.

At last, the Faster R-CNN detector with the small MobileNetv3 backbone appears to react very differently to the augmentations than the previous two models. In our trials, \texttt{contrast}, \texttt{background}, and \texttt{invert} were the most important data augmentations explaining about 32.5\% of the variance.
The least important augmentations according to \acrshort{fANOVA} are \texttt{multiply}, \texttt{median\_blur}, and \texttt{multiply\_brightness}, below 1\% importance.

It is striking how much the importance of data augmentations differ from one model to another. 
One reason is most likely the very different model capacities. Our implementation of \acrshort{yolo}-v7 has more than 8 times the number of trainable parameters than Faster R-CNN MobileNetv3 with about 18 million parameters (see~\cref{tab:training_parameters}).
In our data, only for the large \acrshort{yolo}-v7 the very aggressive space augmentations play an important role for the performance. 
Overall, the data suggests that model capacity correlates with the importance of augmentations that severely change the image or deteriorate information.
Another likely contribution to this is the relatively small number of startup trials. The search space is greater by a few number of magnitude and hence, the initialization plays a pivotal role.
However, there are still some key findings we can derive from the \acrshort{fANOVA} analysis. The augmentations \texttt{invert}, \texttt{simplex\_noise}, \texttt{enhance\_color}, and \texttt{additive\_gaussian\_noise} perform well for all three models. Over all runs and models, \texttt{enhance\_brightness}, \texttt{coarse\_dropout}, and \texttt{linear\_contrast} were the least important augmentations.

\subsubsection{Overall Performance.} 
First of all, the overall necessity for data augmentations in the scope of object detection in space environments is further demonstrated and derived through ~\cref{fig:n_augs_vs_perf}. 
Here, we observe that a detector, trained on synthetic images without any augmentation applied during, achieves very low performance as it is unable to bridge the domain gap. 
Both Mask R-CNN ResNet50 and Faster R-CNN MobielNetv3 overfitted to the training distributions resulting in $0.0$\acrshort{mAP}. 
\acrshort{yolo}-v7 was able to reach about $0.15$\acrshort{mAP} severely impeding the performance of the large model. 
However, the impact of adding only a few augmentations results in a substantial performance boost as measured by \acrshort{mAP}.
Yet, no model with less than 10 augmentation active come close to reaching their full potential.
Furthermore, we derive that a certain number of augmentations, the testing performance starts to diminish, likely due to the loss of information if too many augmentations are added on top of each other.
In our data, this effect is more pronounced for \acrshort{yolo}-v7 than it is for the other detectors. This suggests that the specific choice of data augmentations matter for object detection tasks. 

The performance of the best 10 configurations of Mask R-CNN ResNet50, \acrshort{yolo}-v7, and Faster R-CNN MobileNetv3 can be seen in~\cref{tab:overall_performance}. We evaluate the performance according to two metrics: \acrfull{mAP} and \acrfull{IoU}. The \acrshort{IoU} measures the overlap of the predicted box and the ground truth bounding box. In addition, \acrshort{mAP} is a widely used metric for object detection, for instance in the popular COCO challenge~\cite{Lin2014-qb}. The \acrfull{AP} is defined as the area under the precision-recall curve that is calculated at different \acrshort{IoU} detection thresholds. Following the evaluation of the COCO challenge, we evaluate the \acrshort{AP} for 10 \acrshort{IoU} thresholds $[0.5,...,0.95]$ in steps of $0.05$. Moreover, we report the \acrshort{mAP} for the two parts of the test set \texttt{lightbox} and \texttt{sunlamp}. 

In our experiments, \acrshort{yolo}-v7 performs best reaching a \acrshort{mAP} of $0.932$ and a very strong \acrshort{mAP}@75 of $0.979$. The second best model performs slightly better on the \texttt{lightbox} test data, however in the total \acrshort{mAP} the difference is negligible. For \acrshort{yolo}-v7 the drop between the 10th best and very best model is significant, especially for the \texttt{sunlamp} test data. This suggests the importance of well tuned data augmentations.

The performance of the Mask R-CNN detector is worse. However, considering that the model entails four times less parameters, the performance is very convincing. The \acrshort{mAP}@75 is only slightly worse at $0.977$ and a total \acrshort{mAP} of $0.881$. The results of the ten best configurations are much closer together than they were for \acrshort{yolo} and suggest that the method is less sensitive to the choice of data augmentations. 

At last, the smallest of the models performs significantly worse than the other two methods. Yet, the performance is still very good, considering the small model size and simplicity. The best augmentations configuration reached a \acrshort{mAP} of $0.75$ and \acrshort{mAP}@75 of $0.883$ that is easily good enough for most applications. Interestingly, similar to the other two methods, the best performance for \texttt{sunlamp} and \texttt{lightbox} is always achieved with a different set of augmentations.

\subsubsection{Recommended Set of Augmentations}
We have trained over $400$ models across three different model families, namely Faster R-CNN, Mask R-CNN and \acrshort{yolo}. 
Our experiments suggest that there is no set of augmentations that performs best for all model sizes and methodologies. 
However, we found very good configuration for each model which are summarized in~\cref{tab:overall_performance}. 

For training other models, there are some general recommendation that we can derive from the data we collected. 
Across all models, the augmentations \texttt{invert}, \texttt{additive\_gaussian\_noise}, \texttt{enhance\_color}, \texttt{emboss}, and \texttt{simplex\_noise} were important. 
For each model one blurring augmentations was important such as \texttt{median\_blur} or \texttt{motion\_blur}.
The augmentations derived from~\cite{hendrycks2019benchmarking}, namely \texttt{snow}, \texttt{fog}, \texttt{saturate}, and \texttt{contrast}, were also important for the models and showed great potential as long as the severity is on the low end. 
If the model has a high capacity backbone, the space augmentations \texttt{specular} and \texttt{shadow} boost the performance drastically, which was also shown in~\cite{ulmer20236d}.
At last, even though the results in our study were mixed, we suggest to use augmentations \texttt{background}, \texttt{random\_crop}, and \texttt{affine}.
These augmentations do not contribute to the catastrophic loss of information if too many augmentations are applied and their central contribution is to increase the size of the dataset.
We hypothesize, that one reason for their little importance in our study is that the SPEED+~\cite{Park2021-jr} dataset provides enough training variance in terms of geometric appearance.

\subsubsection{Open-Set 2D Object Detection}
Training any object detector requires a lot of time and resources. For instance, training \acrshort{yolo}-v7 on the SPEED+ dataset takes about 19 hours on a high-end GPU as the NVIDIA A100. 
If the data is not available, a dataset has to be generated that is sufficiently large and realistic. 
Moreover, as we demonstrated in this work, training such models requires careful consideration of hyperparameters and to find a satisfying set likely a multitude of models has to be trained

Motivated by this incredible effort and recent developments in open-set object detections, we conduct an experiment using GroundingDINO~\cite{liu2023grounding} that we briefly described in~\cref{subsubsec:grounding}. Such a model can be used without training and reduces the cost and time required to a minimum. 
In this experiment, we ran one complete evaluation epoch and computed the same performance metrics as in our other experiments. 
In addition, we tested different text prompts as seen in~\cref{tab:grounding}. The best prompt -- "satellite" -- reaches a \acrshort{mAP} of $0.602$ and \acrshort{mAP}@50 of $0.886$. 
In other words, if the correct detection threshold is 0.5, it detects 88.6\% of all samples correctly. GroundingDINO performs better on the \texttt{sunlamp}, which entails strong specularities, than on \texttt{lightbox}, which contains many samples that have very low signal-to-noise ratio. 
However, the model performance is not yet sufficient to deploy such a model to any critical scenario. 
\begin{table*}[htb!]
\caption{Evaluation of the open-set detector GroundingDINO with different text prompts as required input.}
\renewcommand{\arraystretch}{1.3}

\begin{center}
\begin{tabular}{|lrrrrrr|}
\hline
Text Prompt & \rott{$\text{mAP}_{lightbox}$} & \rott{$\text{mAP}_{sunlamp}$} & $\text{IoU}$ &  \rott{$\text{mAP}@50$} &\rott{$\text{mAP}@75$} & $\text{mAP}$  \\
\hline
"shiny rectangle with antennas" & 0.454 & 0.562 & 0.742 & 0.822 & 0.527 & 0.483 \\
"NASA satellite with antennas and solar panels" & 0.509 & 0.567 & 0.763 & 0.865 & 0.581 & 0.523 \\
"spacecraft with solar panels and antennas" & 0.525 & 0.591 & 0.759 & 0.854 & 0.622 & 0.544 \\
"metallic space object with foil, antennas and solar panel" & 0.547 & 0.627 & 0.773 & 0.893 & 0.651 & 0.567 \\
"metallic space object" & 0.542 & 0.637 & 0.775 & 0.862 & 0.645 & 0.567 \\
"satellite" & \textbf{0.583} & \textbf{0.655} & \textbf{0.786} & \textbf{0.886} & \textbf{0.683} & \textbf{0.602} \\
\specialrule{.8pt}{0pt}{2pt}
\end{tabular}
\label{tab:grounding}
\end{center}
\end{table*}

\subsubsection{Inference Time and Memory Consumption}
A very important consideration for orbital perception are the hardware requirements of the target system. 
Mission systems are extremely constrained in terms of computational resources, yet inference time plays a central role due to downstream actions such as robotic grasping. 
To this end, we evaluate the real GPU memory consumption during inference using the \texttt{nvidia-smi} tool. 
In addition, we run inference for 1000 samples and measure the time to process each sample on the NVIDIA Jetson Orin\footnote{https://www.nvidia.com/en-us/autonomous-machines/embedded-systems/jetson-orin/} and a NVIDIA RTX A4000. 
The NVIDIA Jetson board family has recently received a lot of interest for future space missions\footnote{https://www.satellitetoday.com/technology/2024/06/10/planet-to-use-nvidia-ai-processor-onboard-pelican-satellite/}\footnote{https://www.space.com/ai-nvidia-gpu-spacex-launch-transporter-11}. 
Our results are shown in~\cref{tab:hardware_performance}, all values are based on images of size $960 \times 600$. 
\begin{table}[htb!]
\addtolength{\tabcolsep}{-0.3em}
\renewcommand{\arraystretch}{1.3}
\caption{GPU Memory Requirements and Inference Time.}
\begin{center}
\begin{tabular}{|lccc|}
\hline
& VRAM usage & Jetson time & A4000 time\\
\hline
Mask R-CNN & 1.4 GB & 0.5s & 0.09s \\
YOLO-v7 & 1.8 GB & 0.4s & 0.06s \\
Faster R-CNN & 0.6 GB & 0.06s & 0.01s\\
GroundingDino & 1.7 GB & 1.1s & 0.12s \\
\hline
\end{tabular}
\label{tab:hardware_performance}
\end{center}
\end{table}

The Faster R-CNN detector with the MobileNetv3 backbone has the overall smalles memory footprint and inference time that allows object detection above 10Hz on the Jetson. 
The \acrshort{yolo} architecture performs faster than Mask R-CNN, yet requires more GPU memory. 
At last, the large, transformer-based GroundingDINO is the slowest of the methods. 

\section{Conclusion}
In this work, we have shown the importance of data augmentations to bridge the domain gap between synthetic and real-world images in the scope of computer vision applications in orbit. 
To this end, we investigated the impact of applying data augmentations during training on three different networks, Mask R-CNN, Faster R-CNN, and \acrshort{yolo}-v7 on the SPEED+ dataset for satellite object detection.
We have performed a hyperparameter optimization over a set of 29 common augmentations, often used in terrestrial computer vision tasks, with additional, custom space augmentations, simulating common visual effects encountered on objects in space. 
The results of this search is a set of key and performance-boosting augmentation techniques for each detector class.
To spark the use of data augmentation techniques in space, we share these implementation as well as the resulting list with the community.
Additionally, we benchmark GroundingDINO, an open-set detector that does not need to be trained on satellite imagery, on the same real-world test data.
This ablation and poor performance further underlines the unique visual conditions that computer vision models are exposed to in the space domain.
To conclude, we hope that this work contributes to making object detectors, applied to the space domain, more robust and widely used. 

\acknowledgments
The authors gratefully acknowledge the computational and data resources provided through the joint high-performance data analytics (HPDA) project “terrabyte” of the German Aerospace Center (DLR) and the Leibniz Supercomputing Center (LRZ).

\bibliographystyle{IEEEtran}
\bibliography{IEEEabrv,ref}




\thebiography
\begin{biographywithpic}
{Maximilian Ulmer}{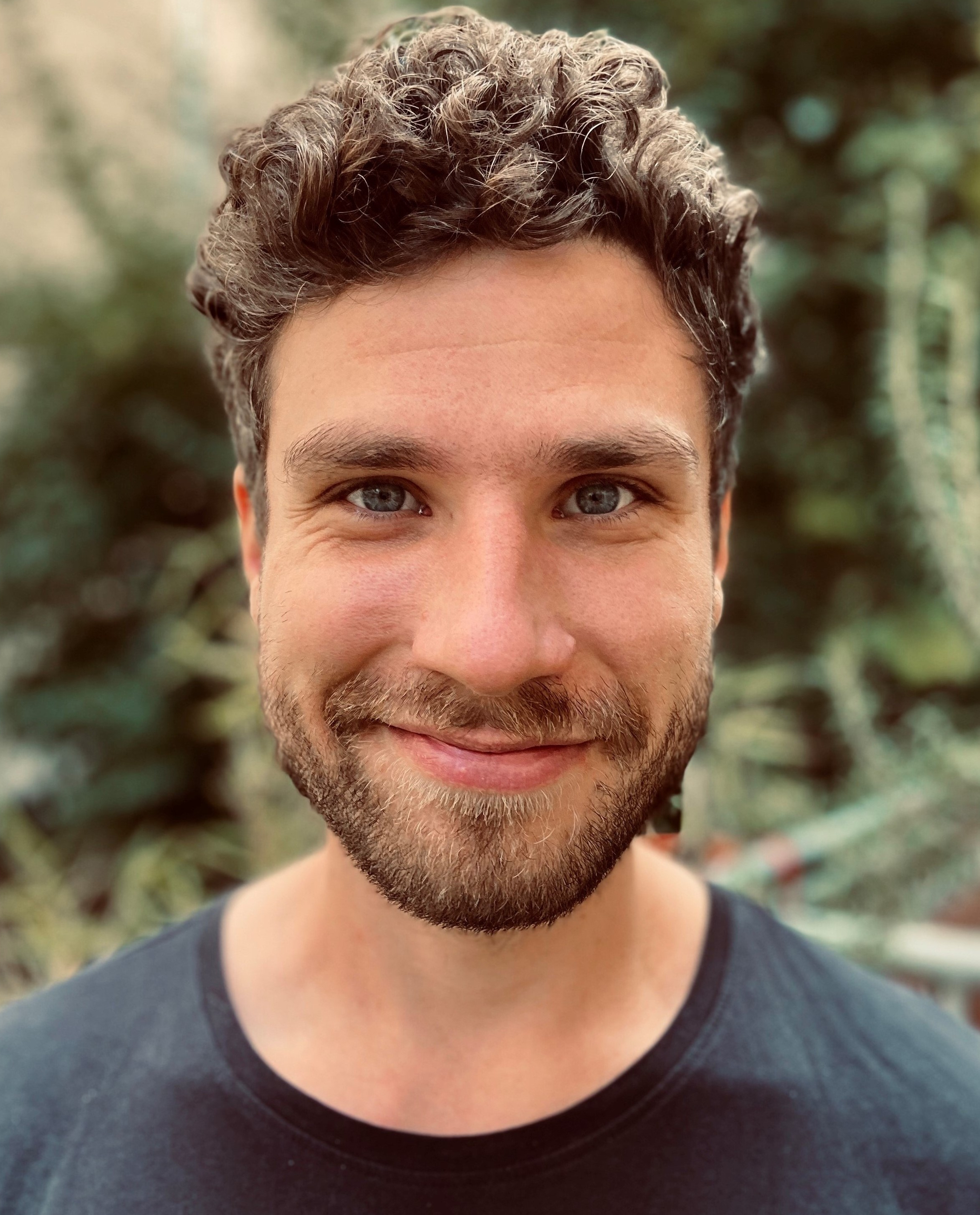}
received his B.Sc.\ and M.Sc.\ in Electrical Engineering and Information Technology from the Technical University of Munich (TUM).  He is currently a Ph.D.\ student at the Karlsruhe Institute of Technology (KIT) and a research scientist at the German Aerospace Center (DLR) in the department for Perception and Cognition at the Institute of Robotics and Mechatronics. His research focuses on closing the Sim2Real gap for orbital perception, 3D vision, and object-centric computer vision for robotic manipulation.
\end{biographywithpic}

\begin{biographywithpic}
{Leonard Klüpfel}{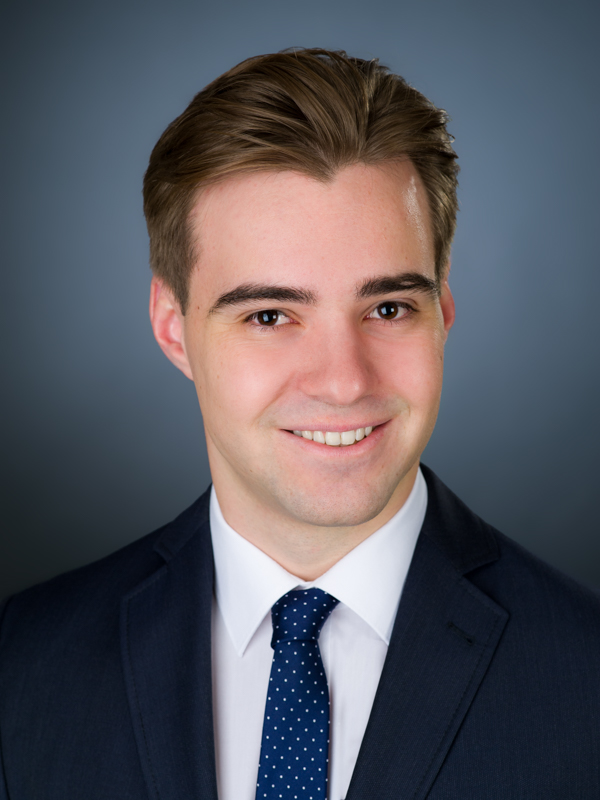}
received his B.Sc.\ in Industrial Engineering from the Friedrich-Alexander University of\ Erlangen-Nürnberg and his M.Sc.\ in Robotics, Cognition, Intelligence from the Technical University of Munich (TUM).  
He is currently a Ph.D.\ student at the Karlsruhe Institute of Technology (KIT) and a research scientist at the German Aerospace Center (DLR) in the department for Perception and Cognition at the Institute of Robotics and Mechatronics. 
His research focuses on computer vision for semantic scene understanding in the context of robotic manipulation with a special interest in detection and tracking of articulated objects with physical considerations and inspiration. 
\end{biographywithpic} 

\begin{biographywithpic}
{Maximilian Durner}{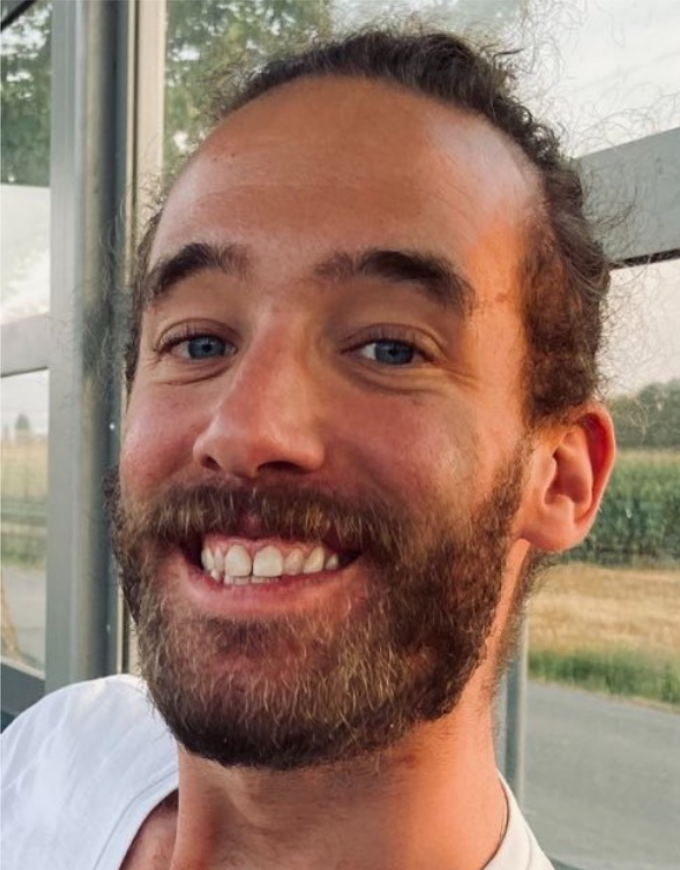}
is a research fellow at the Institute of Robotics and Mechatronics, German Aerospace Center (DLR) since 2016 and a Ph.D. student at the Technical University of Munich (TUM). Before, he studied Electrical Engineering at the TUM, partially studying at the Politecnico di Torino, Italy, and the Universidad Nacional de Bogota, Colombia. Currently, he is leading a research group on semantic scene analysis in the Perception and Cognition Department at DLR, which is working on robotic perception providing semantic information to interact with the surroundings. This includes known, unseen, and unknown object detection, pose estimation, and tracking. In this context, Maximilian focuses on robust and continuously adapting object-centric perception for mobile manipulation.
\end{biographywithpic}

\begin{biographywithpic}
{Rudolph Triebel}{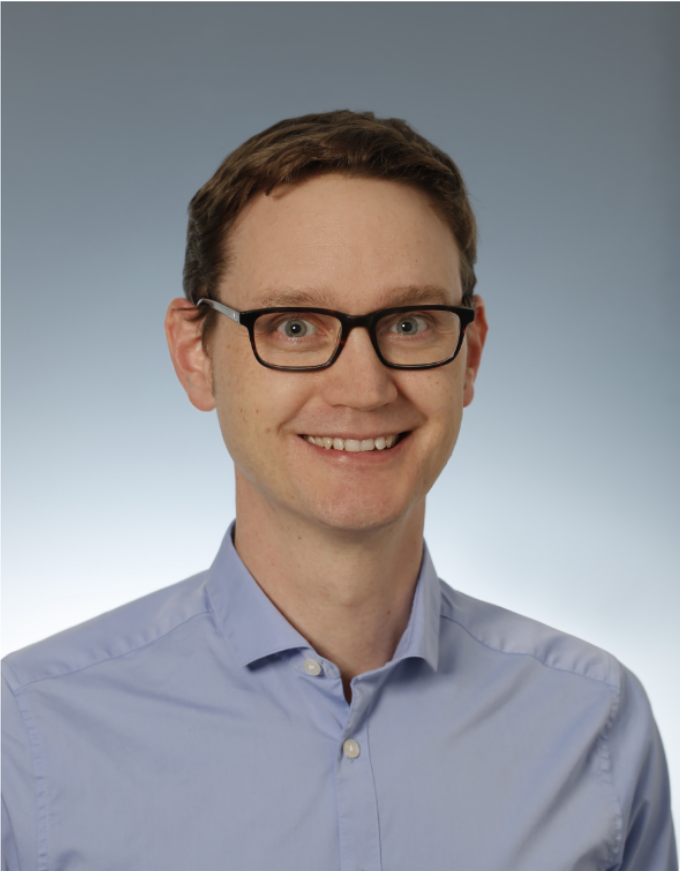}
received his PhD in 2007 from the University of Freiburg in Germany. The title of his PhD thesis is “Three-dimensional Perception for Mobile Robots”. From 2007 to 2011, he was a postdoctoral researcher at ETH Zurich, where he worked on machine learning algorithms for robot perception within several EU-funded projects. Then, from 2011 to 2013 he worked in the Mobile Robotics Group at the University of Oxford, where he developed unsupervised and online learning techniques for detection and classification applications in mobile robotics and autonomous driving. From 2013 to 2023, Rudolph worked as a lecturer at TU Munich, where he teached master level courses in the area of Machine Learning for Computer Vision. In 2015, he was appointed as leader of the Department of Perception and Cognition at the Robotics Institute of DLR, and in 2023 he was appointed as a university professor at Karlsruhe Institute of Technology (KIT) in "Intelligent Robot Perception".
\end{biographywithpic}

\end{document}